# A Graph-Matching Approach for Cross-view Registration of Over-view and Street-view based Point Clouds

Xiao Ling[a,b], Rongjun Qin[a,b,c,d,*]

[a]*Geospatial Data Analytics Laboratory, The Ohio State University, 218B Bolz Hall, 2036 Neil Avenue, Columbus, OH 43210, USA.*
[b]*Department of Civil, Environmental and Geodetic Engineering, The Ohio State University, 218B Bolz Hall, 2036 Neil Avenue, Columbus, OH 43210, USA.*
[c]*Department of Electrical and Computer Engineering, The Ohio State University, 205 Dreese Lab, 2036 Neil Avenue, Columbus, OH 43210, USA.*
[d]*Translational Data Analytics Institute, The Ohio State University.*
*\*corresponding author: Tel: +1 614 2924356, Email address: qin.324@osu.edu (Rongjun Qin)*

**Abstract**

Wide-area 3D data generation for complex urban environments often needs to leverage a mixed use of data collected from both air and ground platforms, such as from aerial surveys, satellite, and mobile vehicles, etc. On one hand, such kind of data with information from drastically different views (ca. 90° and more) forming cross-view data, which due to very limited overlapping region caused by the drastically different line of sight of the sensors, is difficult to be registered without significant manual efforts. On the other hand, the registration of such data often suffers from non-rigid distortion of the street-view data (e.g., non-rigid trajectory drift), which cannot be simply rectified by a similarity transformation. In this paper, based on the assumption that the object boundaries (e.g., buildings) from the over-view data should coincide with footprints of façade 3D points generated from street-view photogrammetric images, we aim to address this problem by proposing a fully automated geo-registration method for cross-view data, which utilizes semantically segmented object boundaries as view-invariant features under a global optimization framework through graph-matching: taking the over-view point clouds generated from stereo/multi-stereo satellite images and the street-view point clouds generated from monocular video images as the inputs, the proposed method models segments of buildings as nodes of graphs, both detected from the satellite-based and street-view based point clouds, thus to form the registration as a graph-matching problem to allow non-rigid matches; to enable a robust solution and fully utilize the topological relations between these segments, we propose to address the graph-matching problem on its conjugate graph solved through a belief-propagation algorithm. The matched nodes will be subject to a further optimization to allow precise-registration, followed by a constrained bundle adjustment on the street-view image to keep 2D-3D consistencies, which yields well-registered street-view images and point clouds to the satellite point clouds. Our proposed method assumes no or little prior pose information (e.g. very sparse locations from consumer-grade GPS (global positioning system)) for the street-view data and has been applied to a large cross-view dataset with significant scale difference containing 0.5 m GSD (Ground Sampling Distance) satellite data and 0.005 m GSD street-view data, 1.5 km in length involving 12 GB of data. The experiment shows that the proposed method has achieved promising results (1.27 m accuracy in 3D), evaluated using collected LiDAR point clouds. Furthermore, we included additional experiments to demonstrate that this method can be generalized to process different types of over-view and street-view data sources, e.g., the open street view maps and the semantic labeling maps.

# 1    Introduction

Coupled street-view and over-view images are regarded as a typical cross-view dataset, the perspective view differences of which are ca. 90° and more. Co-registering such dataset is scientifically challenging yet practically very useful, as it allows to spatially reference a large amount and a mixture of heterogeneous data for purposes of 3D modeling, data conflation and geo-referencing. Often the over-view dataset comes with accurate or at least approximated geo-referenced information, and street-view with partial (often inaccurate) or no geo-referencing information but reflect high resolution and detailed information on façades of buildings and other objects, thus being naturally complementary. In addition, over-view datasets, particularly those from high-resolution satellite sensors, are consistently injecting contents to data archives with global coverage, and street-view datasets can be nowadays more cheaply captured with mobile agents running along the streets (mobile vehicles, crowd-sourcing videos and images), which may lead to much lower cost when these data are being used. Having data from both views geometrically consistent will greatly benefit 3D data collection applications as it provides an alternative solution in cases where high quality data (e.g. decimeter or centimeter level oblique imageries, LiDAR (Light detection and ranging) point clouds) are not available or too expensive to acquire, and cases that data scanned from drastically different views need to be combined for practical applications such as for full three dimensional modeling for smart cities (Heinly et al., 2015).

Turning well-captured over-view images (e.g. photogrammetric image blocks) into 3D geometry are nowadays standard practices for typical urban and suburban areas, which yields reasonably accurate 3D point clouds or meshes that present the scene (Colomina and Molina, 2014). However, cases for processing street-view images are more variable with respect to cameras to be used (e.g. low-cost camera with large lens distortions), disturbances from moving objects, convergence images and scene complexities, thus often yield complicated and non-rigid distortions (Remondino et al., 2017; Wu, 2014). For example, a typical case is the use of single-trajectory video frames for 3D street-view scene reconstruction, which theoretically presents sub-optimal camera networks which might lead to topographical distortion (Dall'Asta et al., 2015) and trajectory drifts (Nobre et al., 2017) in the resulting geometry. Therefore, because 3D geometry generated from the over-view datasets is more metrically correct, it can serve as the base dataset to which 3D geometry generated from the street-view can align.

Although there have been attempts to address similar problems such as cross-view localization (Castaldo et al., 2015; Hu et al., 2018; Liu and Li, 2019) (i.e. localization on an over-view image given a ground-view) and cross-view synthesis (Regmi and Borji, 2018; Zhai et al., 2017; Lu et al., 2020; Toker et al., 2021) (simulating over-view using ground-view or vice versa), robust cross-view data co-registration appears to be more challenging and has not yet been well investigated. In general, co-registration of such over-view and street-view data encounters three major challenges:

1. Almost all the sensors collecting the 3D information are bearing-only and produce information only at line of sight, therefore the over-view and street-view dataset, due to the significant view differences, share very limited areas in common for extracting textural or geometric correspondences to build observations (Shan et al., 2014);
2. For cases that no georeferencing information is available, searching for coarse positional alignment to initiate the co-registration can be extremely difficult given the complexity of geometry in the street view;
3. The non-rigid topographical distortions often presented in street-view 3D geometry generated from a single-trajectory data collection hinder the use of simple transformation models such as a similarity or rigid transformation, in which more complicated and potentially non-parametric models are needed (Bruno and Roncella, 2019; Lee, 2009).

These three challenges are essentially difficult cases in which common solutions used for 3D data co-registration often failed (Cheng et al., 2018; Pomerleau et al., 2015). For example, for challenge 1), none of the existing image feature extraction and matching methods are capable of processing cross-view images without overlapping texture (Morel and Yu, 2009) and 3D feature extraction & matching method are rather immature even for well-captured and high-quality 3D data (Hana et al., 2018). Although there exists learning based methods that learn feature correspondences in a cross-view scenario (Hu et al., 2018; Tian et al., 2017), such methods are usually sensitive to availability of training samples and are often subjected to a lack of transferability and generalizability for regions that are geographically and contextually different (Bengio et al., 2013). Solutions for challenge 2) often based on local and structural information to build correspondences; a typical application scenario is on navigation in GPS-denied environments, in which one needs to incrementally evaluate the built trajectory from video frames and map overlays, and formulate the matching through graphs to build approximate correspondences (Mourikis and Roumeliotis, 2007; Vaca-Castano et al., 2012). For challenge 3), although there exist a number of approaches for co-registration of deformable 3D shapes (Papazov and Burschka, 2011), these methods mostly focus on fitting pre-existing parameterized models (e.g. human body with underlying skeleton models) (Marin et al., 2020). However, the case of topological distortion in our context, is as a result of inaccurate camera poses, interior orientations and lens distortions, which may vary with the terrain and cannot be simply addressed using parametric models for correction.

The intent of this work is to provide an accurate solution to co-register the street-view and over-view 3D data in a large urban region, in which we assume very little or no geo-referencing information for street-view data. Given that the over-view and street-view data are collected in an approximated 90◦ view difference, it is reasonable to assume that the object boundaries (e.g., buildings) from the over-view data should coincide with footprints of façade points from street-view. Therefore, in this paper, we propose to take advantage of the semantic information (e.g., buildings) for cross-view registration. Specifically, we model segments of buildings as nodes of graphs, both detected from the satellite-based and street-view based point clouds, thus to form the registration as a graph-matching problem to allow non-rigid matches; to enable a robust solution and fully utilize the topological relations between these segments, we propose to address the graph-matching problem on its conjugate graph solved through a belief-propagation algorithm. The matched nodes will be subject to a further optimization to gain precise piece-wise smooth rigid 3D transformations, followed by a constrained bundle adjustment on the street-view image to correct non-rigid distortions over the entire trajectory, to yield well-registered street-view images and point clouds at least with the same order of magnitude of the over-view data resolution (or, GSD (ground sampling distance)).

The rest of this paper is organized as follows: **Section 2** presents a literature survey related to relevant methodologies in 2D/3D cross-view registrations. **Section 3** describes the proposed methodology for data preprocessing and cross-view registration. Experimental results and comparative studies on a cross-view dataset consisting of satellite and ground-view images are presented in **Section 4**. **Section 5** concludes this paper by analyzing the advantages and drawbacks of our methodology that inform our planned future works.

## 2  Related work

Algorithms addressing point cloud registration vary with the 3D data including their resolution, level of overlap, accuracy, assumed outlier/error models and heterogeneities between two 3D scans. There are a few widely-practiced algorithms in 3D point clouds registration for generally well-collected 3D point clouds, e.g., those from single-sources, being noiseless, and with sufficient overlaps (at least 20%) (Stechschulte et al., 2019), such as the Iterative Closest Point (ICP) methods (Besl and McKay, 1992; Zhang, 1994), its variants (e.g. GoICP(Global optimal ICP) (J. Yang et al., 2015), symmetric ICP (Rusinkiewicz, 2019), fast robust ICP (Zhang et al., 2021)) and classic least-squares surface matching (Gruen and Akca, 2005). In particular, the ICP algorithm (and potentially its variants) is widely implemented into open-source / commercial software packages (Girardeau-Montaut, 2020; Rusu and Cousins, 2011), that often serve as the "first-trial" baseline when attempting a point cloud registration problem. The ICP algorithm and its variants assume the point clouds to first start with good initial positions (coarsely aligned), usually guided by manually collected 3D correspondences or external observations such as from GPS/IMU (Inertial Measurement Unit), and then the algorithm iteratively optimize transformation parameters (rigid or affine) by minimizing sum of point-wise distance metrics between assumed correspondences. Research focuses on these topics are to achieve better convergences of optimization given coarse initializations, for example, the GoICP (J. Yang et al., 2015) algorithm is capable of finding a global solution for point cloud registration, which as an improved version of the classic ICP method, utilized the BnB search (Breuel, 2003) to step out of local minimum that ICP trapped in using an iterative solution, while the work only assumed a rigid transformation while leaving more complicated models undiscussed. Besides the effort in achieving more robust and accurate matching with two pieces of 3D point clouds, only a few studies have been published for cases where point clouds are scanned from drastically different views and with potentially different resolution, the level of occlusions and lack of overlap easily fail foregoing algorithms, an example is the work proposed by B. Yang et al. (2015), which extracted building outlines from airborne and terrestrial laser scanning point clouds respectively and then applied a general graph-matching approach specifically targeted on these outlines to obtain building correspondences, however it relied on well-delineated and complete line features which are luxury for noisy data like us. Furthermore, if any of the 3D point clouds present non-rigid distortions (e.g. due to systematic errors introduced by data capture, such as point clouds generated by images captured by uncalibrated data), these type of well-defined registration algorithms assuming parametric transformation models may not work.

The classic 3D registration solutions (e.g. ICP) assume structured dense correspondences (i.e. one-to-one correspondence for all points in the overlapping region), while to accommodate more general scenario where 3D pieces are incomplete and noisy, solutions are sought by using pattern recognition techniques adapted to 3D structures for interest feature detection and matching, with the goal to independently locate 3D corresponding points that are distinctive and with high confidence. Popular 3D features include 3D points (Castellani et al., 2008; Sipiran and Bustos, 2011) and 3D line segments (Lin et al., 2017). The identification and matching of the 3D interest points are based on either handcraft features descriptors such as spin image (Johnson and Hebert, 1999), shape context (Belongie et al., 2000), and distribution histogram (Anguelov et al., 2005), or learnable descriptors such as local volumetric patch descriptor (Zeng et al., 2017), fully-convolutional geometric features (FCGF) (Choy et al., 2019) and combined multi-Layer Perception (CMLP) (Huang et al., 2020). However, these descriptors of 3D key points and lines utilize local structures of the 3D data, which can hardly be built equivalent on scans of 3D data coming from completely different views, and even more challenging when there exist large resolution difference and very little overlap between scans.

Non-rigid and unknown distortions of the point clouds present as the largest hurdle when performing registration between two 3D point clouds scans, since on one hand, selecting the appropriate transformation model can be challenging, and on the other hand, the optimization of transformation parameters may get more difficult as the model is getting more complex. Existing solutions assume simple shaped models, by considering this unknown transformation to be approached by using mixture models, such as the work of Ma et al. (2018); Myronenko and Song (2010), where the potential over-parametrization was addressed by incorporating manifold regularization through the well-known Expectation Maximization (EM) algorithm (Myronenko and Song, 2010); a similar attempt solved these non-rigid parameters through kernel correlation maximization problems through Gaussian Mixture models (Jian and Vemuri, 2010). However, these methods worked with assumed non-rigid transformation model and compact and simple shaped geometry with good overlaps, thus can hardly be applied directly to cross-view registration cases where non-rigid distortion are more challenging. For point clouds generated from monocular video frames collected from a moving platform, as to be in this work, the resulting long and thin point clouds along the street may subject to severe trajectory drift owing to inaccurate camera interior orientations and the suboptimal camera networks, thereby the geometry may be inconsistently piecewise rigid/non-rigid, and cannot be simply modeled using a few transformation parameters.

To sum, these multiple challenges, i.e. the lack of overlap for reliable 3D features, drastically view difference, and non-parametric distortions of geometry, have positioned the cross-view point cloud registration a unique problem to solve. Despite there are a few efforts designated to address the "cross-view" aspects by using deep learning methods to correlate ground-level and over-view images (Liu and Li, 2019; Hu et al., 2018; Tian et al., 2017; Castaldo et al., 2015), these work mainly focused on the "localization" aspects that meant to provide an image-level correspondence and so far were incapable of achieving pixel/point level correspondence identification cross different views. This paper aims to address the challenges specifically related to cross-view point cloud registration by 1) taking semantic object (e.g. buildings) boundaries as view-invariant feature for street-view to over-view matching, and 2) converting the global registration into a graph-matching and energy minimization problem and solved by belief propagation to obtain segment-level correspondences for registration. The rationale for using building boundary is that the boundaries from the over-view data should coincide with footprints of façade points from street-view, and the theoretical registration accuracy of building boundaries should be in the same magnitude as the GSD of the overview satellite images, which should be much more accurate than start-of-the-art image localization methods. Moreover, the using of graph matching can largely reduce the search space for the non-rigid registration, which enables the global solution for registering the cross-view point clouds with arbitrary initialization. Finally, the segment-level registration matches the assumption of piecewise rigid transformation well and can be performed robustly and efficiently.

### 3      Methodology

Given a point cloud generated by over-view sources and a point cloud collected from the street-view images by a monocular camera (with very limited or no geo-referencing information), we define the cross-view 3D point cloud registration as to precisely register the street-view point clouds to the over-view point clouds at the point level. Our solution, as mentioned above, relies on object boundaries casted on the ground, and performs segment-level matching through a global optimization to address piecewise rigid transformation, followed by point-cloud level fine registration. **Figure 1** presents the general workflow of the proposed co-registration method: starting from producing the point clouds from both the over-view source (here we used multi-view satellite images at 0.5 GSD) and the street-view images (video frames from a single GoPro camera) as a preprocessing step; in a second step, semantic information (i.e., building segments) are extracted from both street-view and over-view point clouds using methods in **Subsection 3.2**. With the extracted buildings from both views, the segment-level correspondences are then formulated as a graph-matching problem solved by belief-propagation to yield globally consistent 2D segment matches, followed by point cloud level fine registration. After the 3D registration, we further apply a constrained bundle adjustment on the street-view images to enable the image poses to be consistent with the registrations as well, with which the dense point clouds can be optionally reproduced. Details of those steps are introduced in the following sub-sections.

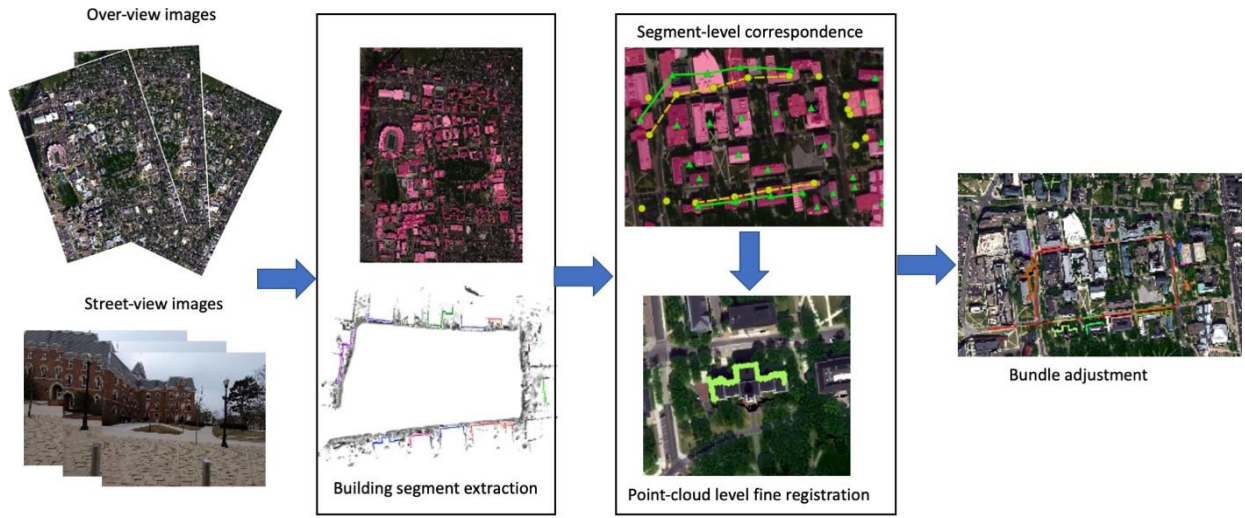

*Figure 1*: Workflow of the proposed cross-view registration algorithm. Details of each component are introduced in the subsections.

## 3.1 Preprocessing

The input over-view images consist of 0.5m resolution Worldview (I/II) images, and the street-view images are collected by a single GoPro video camera (more details about the data will be introduced in **Section 4**). The preprocessing for separately converting the over-view and street-view images follow standard and existing photogrammetry and dense matching techniques: The multi-view satellite images are processed following a multi-stereo approach as described in (Qin, 2017, 2016), which performs a pair-wise reconstruction to DSM, followed by a DSM (Digital Surface Model) fusion. The core matching algorithm uses a hierarchical Semi-Global Matching (Hirschmuller, 2008) with modifications to accommodate large-format images (Qin, 2014), and the approach for selecting and ranking the pairs follows the approach as described in (Qin, 2019) based on the available images and their metadata (DigitalGlobe, 2020).

The street-view video frames are processed through an enhanced structure from motion (SfM)/photogrammetry approach which introduces a few strategies from the SLAM (Simultaneous Localization and Mapping) community (Mur-Artal and Tardós, 2017), i.e. with velocity models for feature tracking to enable robust incremental relative orientation and bundle adjustment, and it should be noted that without these strategies, a standard SfM may unlikely to succeed for a longer trajectory of video frames due to the unstable feature correspondences detected merely by standard feature operators (Lowe, 2004). Dense matching over these oriented frames is performed using a multi-view approach implemented in the open-source software OpenMVS (Cernea, 2015), the results of which are shown in multiple figures in this paper (e.g. **Figure 3(a), Figure 7(c), Figure 11(a)** etc.). Note in our work, we assume the scale of the model is known based on measurements of known dimensional objects or coarse GPS information.

## 3.2 Building segment extraction

We assume the registration model from the street-view point clouds to the over-view data to be piece-wise rigid, meaning that registration from a section of the street-view point clouds to the over-view data follows a rigid transformation, while different sections of the point clouds might not share the same rigid transformations. Therefore, the idea here is to firstly segment both the over-view and street-view point clouds, and potentially correspond them to facilitate matching to accommodate our assumption on the piece-wise rigid transformation model. Therefore, in a first step, segments from the buildings or any off-terrain object (since they are primarily buildings, we call them building segments for simplicity hereafter) need to be extracted for segment-level matching.

*__Building segment extraction from over-view data.__* To take the boundary of buildings as the cue for registration, we first extract building segments from the over-view data. Here we assume orthophoto and DSM are available (can be easily converted from the dense point clouds if not). The extraction of building segments in our experiments follows a heuristic approach based on a grey-level top-hat based detector (Vincent, 1993; Qin and Fang, 2014) on the DSM, followed by a few cascade filtering using indices such as NDVI (normalized difference vegetation index) (Carlson and Ripley, 1997) and shadow index (Huang and Zhang, 2011). This extracts a building mask and individual building segments are extracted using a connected component analysis (Grana et al., 2010). An example of results is shown in **Figure 2**, and it can be seen that artifacts exist such as those detected segments in the river, and our approach in matching these segments necessarily consider such erroneous detections (introduced in **Section 3.3**). It should be noted that the building segment extraction algorithm is replaceable by other and possibly more advanced algorithms to obtain more accurate building segments, such as deep learning based detectors (Guo et al., 2020; Cheng and Fu, 2020), or by any existing and accurate building footprint data (e.g. some high-quality OpenStreetMap data (OpenStreetMap, 2021)) if well-aligned with the over-view images. Note when 3D data are not available, our approach may be able to register the street-view point clouds to be horizontally aligned with the building footprints, given the nature of our method intending to align the street-view point clouds to the over-view data through the building boundaries.

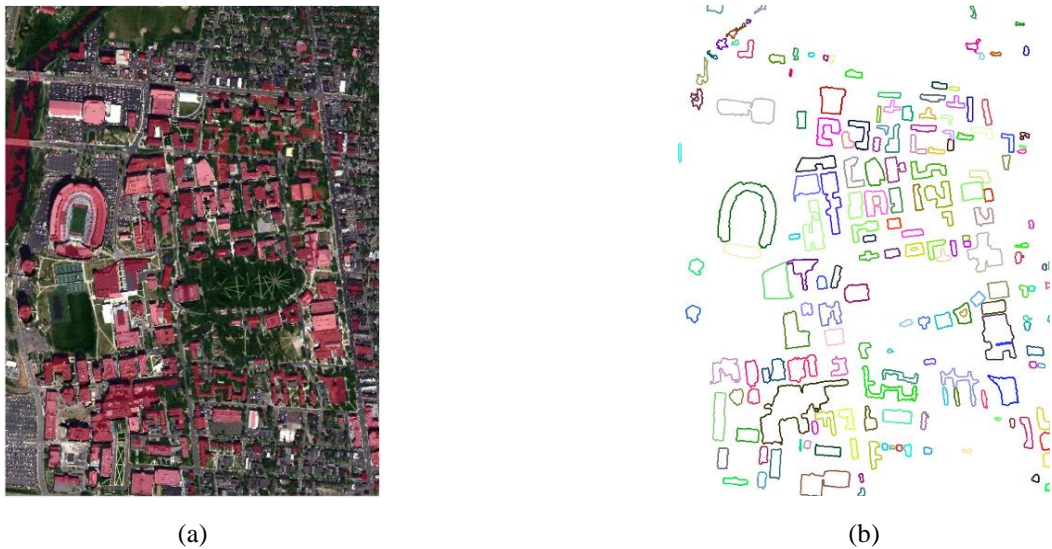

(a)          (b)

*Figure 2*: An example of building segment extraction from the over-view multispectral orthophoto and DSM. (a) The orthophoto with buildings highlighted in red, (b) Individual building segments shown in different colors.

*__Building segment extraction from street-view data.__* To identify building segments from the street-view point clouds, we first separate façade points from the ground points based on their normals. Here we assume that the Z plane of the street-view point clouds approximately aligns with the ground coordinate frame, thereby points with normals orthogonal to the vertical direction can be identified as façade points. To estimate the point-wise normal, we built a KD-tree structure (Muja and Lowe, 2014) for each point $p_i$ to find their $k$-nearest-neighboring point set $N_i$, which will be used to calculate its normal vector $v_i$. A histogram of orientations of all points are built and the most frequent direction is further validated and determined as the vertical direction $v_v$, and this will allow a correction of the $v_v$ to be aligned with the vertical direction of the satellite point clouds (whose vertical direction can be extracted following the same method) through a rotation applied to the street-view point clouds. Points with their normal direction approximately perpendicular to the vertical direction $v_v$ are considered to be façade points. We specifically define the following two criteria to be met for determining a point $p_i$ with normal direction $v_i$ as a façade point: 1) the intersection angle between $v_i$ and $v_v$ is bigger than 75 degrees; 2) more than 75% neighbors in $N_i$ meet 1) to robustify the decision.

In practice we find this process to be very effective in determining façade points, which produces primarily building façade point clouds with a few other small objects such as trees. We apply a standard region-growing segmentation (Rabbani et al., 2006) method to identify individual building segment as well as eliminating small segments of points, here in this work we eliminate any point clusters whose lengths are less than 5m in diameter horizontally. Figure 3 shows the extracted point clouds segments and it can be observed that almost all of these segments remains to be part of man-made/building objects. These dense points can be then further projected onto the Z plane to yield 2D building boundaries. It should be noted that Figure 3(a) shows a street-view point clouds of a closed loop, while the errors caused by trajectory drift lead to non-rigid distortions of the whole point cloud. Moreover, although façade points and their projected segments in 2D are extracted and used to serve the 2D building segment registration, the ground points are still associated with the nearest building segments, thus for any subsequent transformation, these points will be transformed with along with the building segment they belong to.

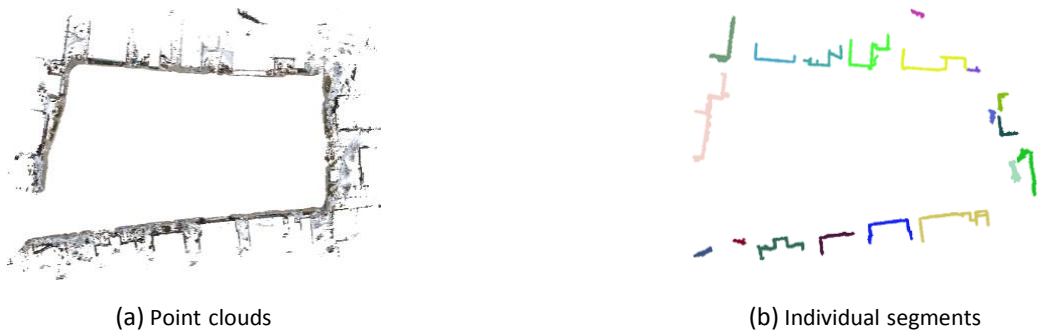

(a) Point clouds          (b) Individual segments

*Figure 3*: Extracted façade segment (b) from the street-view point clouds (a). These segments are colorized to indicate their independence, and each segment will be served as the candidates to match with the over-view building segments. Note the point cloud shows street sides of a closed loop while is open due to error caused by trajectory drifts (non-rigid and non-parametric distortion).

### 3.3 Cross-view registration

With the preprocessing steps mentioned in previous sections (**Section 3.1-3.2**), complete building segments from the over-view data (**Figure 2(b)**) and partial building segments from the street-view point clouds (**Figure 3(b)**) projected to the Z-plane serve as the objects of interest for matching, in which the over-view building segments follow the geodetic coordinate frame derived from the satellite images and the street-view building segments are in arbitrary coordinate frame (with the vertical direction aligned with that of the satellite point clouds and the scale known). Given the non-rigid distortion of the street-view point clouds as shown in **Figure 3(a)**, there is obviously not a single rigid transformation that transforms these 2D street-view building segments to be aligned with the over-view building segments, and with our piece-wise rigid assumption, the first task is to correspond the street-view building segments to the over-view segments (segment-level correspondence), and then perform rigid registrations per segment, with smoothness constraints over the individual rigid model to avoid the registration being too aggressive at a single segment-pair (due to errors), thus to yield global and point cloud level registration.

### 3.3.1. *Segment-level correspondence through graph-matching*

To correspond the 2D segments derived from the over-view and street-view data, the approach must abide by the following possible challenges: 1) there may not exist an initial correspondence; 2) the street-view segments are partial in shape and features can be hardly extracted to correspond to the complete shaped over-view building segments; 3) there might be one-to-many, or many-to-one correspondences or no correspondences. We consider this correspondence problem as graph-matching task, meaning to assign for each street-view segment, an over-view segment as the correspondence. However, given the first two challenges, extracting information that represent distinctive features for matching can be hardly achievable. Therefore, our solution seeks for the use of topological relationships between neighboring segments as a relatively more robust metric to corresponding segments. A concept is shown in **Figure 4**, where pairs of neighboring segments tend to provide more information to match incomplete street-view segments to the over-view segments. Therefore, we consider formulating the problem of matching building segments (segment as a node) into a conjugate graph (an edge linking a pair of segments as the node), which has more information to explore per node, such as orientations of the edge, length of the edge etc. As a result, our solution performs a graph-matching on this conjugate graph, that for each street-view edge $(S_j, S_k)$ (defined as a pair of street-view segments $S_j$ and $S_k$, where $j, k$ are segment indices), we seek for its corresponding over-view edge $(O_p, O_q)$ (defined as a pair of over-view segments $O_p$ and $O_q$, where $p, q$ are segment indices), where the label space is effectively represented as $L = \{e_{p1,q1}, e_{p2,q2}, \cdots, e_{pm,qm}\}$ and $e_{pi,qi}$ refers to an over-view edge (a node in the conjugated graph). Note that in this conjugate graph, although the number of nodes increases as compared to the original graph (one segment per node), we only enable nodes representing a pair of neighboring segments (through $K$-nearest criterion, here we use $K = 4$), thus to greatly reduce the complexity of ($n(n - 1)$ reduced to $4n$, $n$ refers to the number of over-view segments).

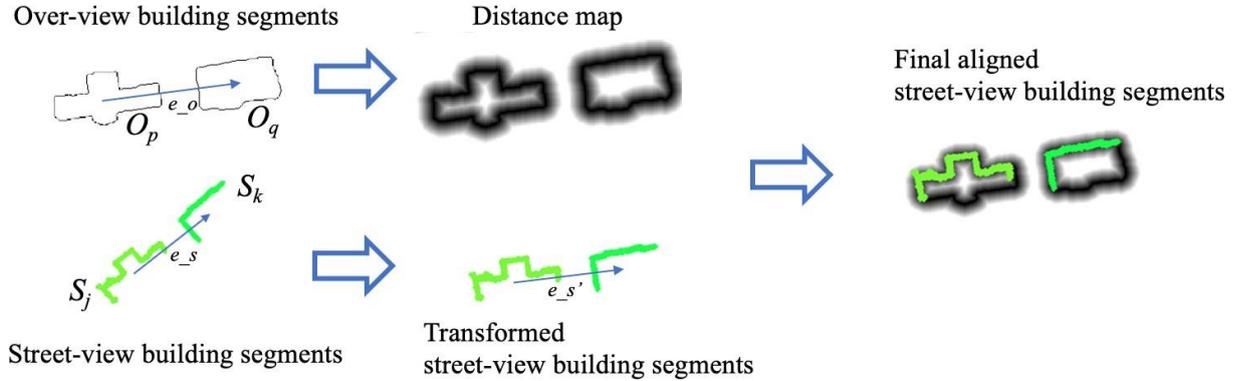

*Figure 4*: A pair of segments serving as a node in the conjugate graph. Given two over-view building segments $e\_o = \{O_p, O_q\}$ and two street-view building segments $e\_s = \{S_j, S_k\}$, the coarse aligned street-view building segments $e\_s'$ is firstly obtained by transforming centers of $e\_s$ to centers of $e\_o$ and the distance map for over-view segments is calculated where the intensity of pixel denotes the distance to closest point in segments, then the final transformation is obtained by moving the $e\_s'$ back and forth to achieve the minimal distance.

Given a street-view edge (a pair of street-view segments) $e\_s$, we evaluate the potential matchiness of a candidate over-view edge $e\_o$ (a pair of over-view segments) by computing their minimal Euclidean distance at the 2D level. This is performed following three steps, first, we abstract both of the edges as lines ($l\_s$ and $l\_o$), the end points of which are centers of the segments, and compute a rigid 2D transformation $T_{s\_2\_o}$ transforming $l\_s$ to $l\_o$; second, apply this 2D transformation to the street-view edge $e\_s$ (or the pair of street-view segments), yielding $e\_s' = T_{s\_2\_o}(e\_s)$ to be approximately aligned with candidate over-view edge $e\_o$. Third, the distance between the $e\_s'$ and $e\_o$, since they are close enough in the 2D plane, can be computed with the aid of distance map (Fabbri et al., 2008): **Figure 4** as an example of the distance map that used to compute 2D shape matches. Here we offset $e\_s'$ within a certain window (i.e. the bounding box of $e\_o$) and take the local minimal of its distance to $e\_o$ as the matchiness of the shapes (denoted as $D$). The offset that achieves the minimal distance can be incorporated into $T_{s\_2\_o}$, denoted as $\widehat{T_{s\_2\_o}}$.

Apparently a "winner-takes-all" strategy based merely on the potential of matchiness between the street-view and over-view edge will be unlikely to yield good registration results, as the quality of segments vary, and such a local solution will generate noisy results. We consider a smoothness constraint that enforces the transformations $\widehat{T_{s\_2\_o}}$ obtained at the segment-level to be consistent for neighboring edges (or neighboring nodes in the conjugate graph), thereby to formulate an energy minimization problem for the conjugated graph that considers the matchiness $D$ described above as the data terms and the smoothness constraint as the smooth term in below:

$$E(\mathbf{L}) = \sum_i^n D(l_i) + \sum_i^n \sum_{j \in \mathcal{N}_i} V(l_i, l_j) \qquad (1)$$

where $D(l_i)$ refers to the data term, being the matchiness computed as described above for a node (a street-view edge) and the over-view edge $l_i \in L = \{e_{p1,q1}, e_{p2,q2}, \cdots, e_{pm,qm}\}$ as the potential candidate. $\mathcal{N}_i$ denotes the neighborhood of $l_i$, here defined as the over-view edges that share a node with $l_i$. $V(l_i, l_j)$, refers to the smooth term, that punishes inconsistent transformation between neighboring nodes (over-view edges). Details about these terms are further given in below.

1. Data term: the costs for each node are normalized to $[0, 1]$; note here we define a null label $l_0$, to represent the case that when a street-view edge does not correspond to any over-view edge, in this case the cost $D(l_0)$ is directly set to 1.

2. Smooth term $V(l_i, l_j)$ is defined in a truncated form that separately evaluating the consistency of rotation and translations from $\widehat{T_{s\_2\_o}}$ for neighboring nodes:

$$V(l_i, l_j) = \begin{cases} c1, & \text{if } (l_i, l_j) \in \mathbb{N}(\mathbf{L}), \text{and } \|\theta_i - \theta_j\| < \theta_{th} \text{ and } \|t_i - t_j\| < t_{th} \\ c1, & \text{if } l_i = l_0 \text{ or } l_j = l_0 \\ c2, & \text{otherwise} \end{cases} \qquad (2)$$

where $\mathbb{N}(\mathbf{L})$ is defined as a neighborhood set that contains over-view edges that shares a common node. Here $(l_i, l_j) \in \mathbb{N}(\mathbf{L})$ means that the smooth term favors (i.e., with a low cost) neighboring street-view edges to have neighboring over-view edges as correspondences. $\theta$ is the rotation angle in 2D, and $t$ is the translation derived from $\widehat{T_{s\_2\_o}}$, $\theta_{th}$ is the angle threshold which is set to 10°, and $t_{th}$ is the translation threshold set to 100 meters. The second condition $l_i = l_0$ or $l_j = l_0$ means the neighborhood of a street-view edge is allowed to not correspond to any over-view edge. $c1$ is set as a smaller value to favor consistent transformation between neighboring edges, otherwise $c2$ (as a bigger value).

Note this energy minimization problem assumes the conjugate graph a Markov-random field (Li, 1994), while the marginal distribution of the statistical variables of the graph (if considering the label for each node) is not independent. Thus, traditional graph-cut solver (Kolmogorov and Zabin, 2004) that aims to maximize marginal distribution will unlikely work. Thus, we use

Preprint of the accepted version, with only minor editorial differences from the final versionthe loopy belief propagation algorithm (Coughlan and Ferreira, 2002), which iteratively minimize the energy through message passing and inference: given a pair of neighboring nodes (i.e. two street-view edges $i$ and $j$ that shares a common node), the directed messages $m_{i,j}$ from $i$ to $j$, which is initialized to 1, is updated by considering all message flowing into $i$ (except for message from $j$) via **Equation 3**, where $\mathbb{N}(i)\setminus j$ is the set of neighboring street-view edges except j for $i$, the cost of each label (over-view edge) $l_i$ for $i$ are computed from **Equation 4** and the smallest cost is taken as its selected label, this procedure iterates until the computed energy **Equation 1** converges, and labels with the smallest cost for each street-view edges are taken as final solutions.

$$m_{i,j}^{new}(l_j) = \sum_{l_i} D(l_i) V(l_i, l_j) \prod_{k \in \mathbb{N}(i)\setminus j} m_{k,i}^{old}(l_i) \tag{3}$$

$$C(l_i) = D(l_i) \prod_{k \in \mathbb{N}(i)} m_{k,i}(l_i) \tag{4}$$

The graph-matching result of the sample trajectory (**Figure 5(a)**) is shown in **Figure 5(b),** where the street-view segments (colored in isolated solid line) are visualized on the top-view building segments (white mask). Center of gravity for the over-view building segments are in solid green line and that of street-view segments is in blue dash line, which are topologically corresponded well. It should be noted that these are fragmental and noisy street-view segments in the red-rectangle region, which failed to find corresponding over-view segments, which are marked as null correspondences by our algorithm.

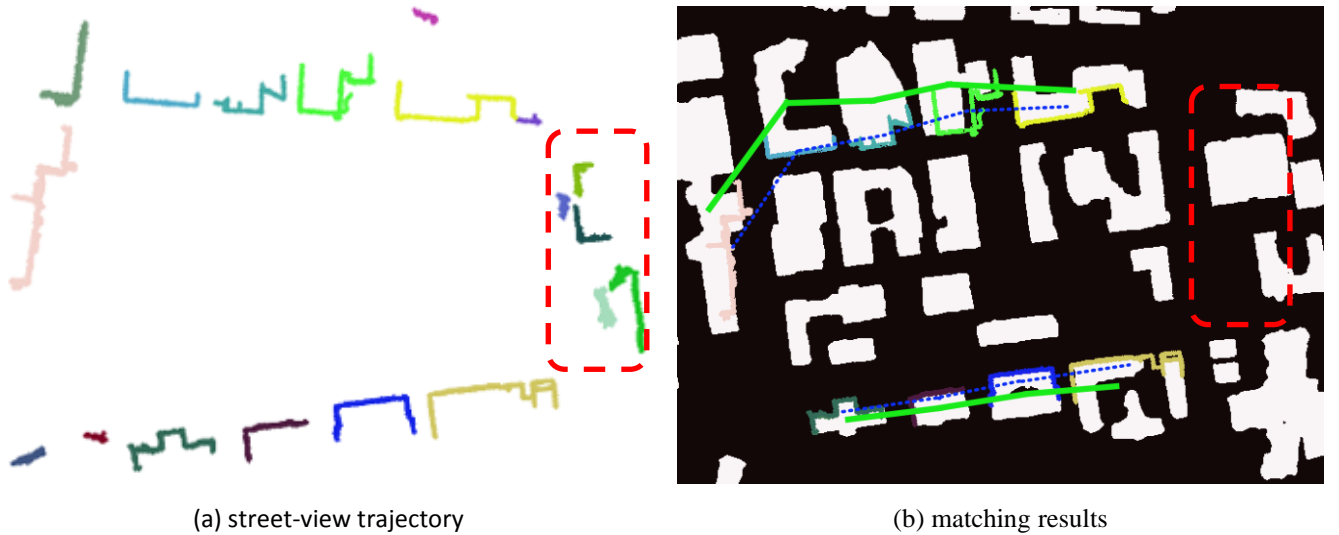

(a) street-view trajectory        (b) matching results

*Figure 5*: Illustration of the result of the segment-level matching for the sample trajectory. (a) building segments from street-view data; (b) matching results. The solid green line connects the centers of buildings from over-view building segments, while the blue dash line connects the centers of building segments, colorized to differentiate different street-view segments. Street-view segments in the red-rectangle region failed to find corresponding over-view segments, which are marked as null correspondences by our algorithm.

There two tunable but dependent parameters $c1, c2$, the difference of which control the smoothness of the transformation

1111

parameters between neighboring buildings. Therefore we fixed $c2$ to 0.6 (as an empirical mid-value (in a scale of 1)) and test how the algorithm performed according to difference $c1$ on a sample trajectory as shown in **Table 1**. It shows different c1, when set between 0.1 and 0.5 will lead to convergence with differed iteration number. According to the table we thus take $c1 = 0.1, c2 = 0.6$ as a good empirical value throughout the experiment.

**Table 1.** Iteration numbers used in graph matching when $c2 = 0.6$ on a sample trajectory.

| $c1$ | 0 | 0.1 | 0.2 | 0.3 | 0.4 | 0.5 | 0.6 |
|---|---|---|---|---|---|---|---|
| Iteration number | diverged | 8 | 7 | 10 | 16 | 42 | diverged |

*3.3.2. Point-cloud level fine registration*

The algorithm described in **Section 3.3.1** provides a solution for matching 2D building segment, here we perform a fine registration at the building segment level by 1) further refining 2D segment-level alignment, and 2) performing a 3D rotation to align the street-view point clouds to the over-view point clouds, and both of these two approaches are rather heuristic:

<u>*Fine 2D registration at the building segment level:*</u> We further refine the 2D transformation parameters by performing an fine-level exhaustive search, where the range of the search for the 2D rotation stays within $[-10°, 10°]$ with a predefined interval (e.g. 1°) as the searching step, and the offset range is within the bounding box of the over-view segments, with one pixel per step. The process can be speed up using a dichotomy search, while in practice this 2D level matching is operated in the image grid with the same resolution as the over-view image (i.e. satellite orthophoto in our case), thus can be processed efficiently. This will yield a 2D rotation matrix $R_i^{2d}$ and offset $t_i^{2d}$.

<u>*Fine-level 3D registration*</u>: To advance the registration to 3D, a fine-level registration is performed to align the Z direction of the street-view point clouds to the over-view point clouds. Although the Z-plane of the entire street-view point clouds have been roughly aligned with the over-view point clouds (as introduced **Section 3.2**), these may still exist rotational errors (primarily roll and pitch). We consider the same approach used for the whole street-view point clouds to correct segment level rotation error: for each building segment and its associated ground points, we first separate the ground points based on the normals (those whose intersection angle with the vertical direction $v_v$ is smaller than 15°, followed by a median filtering to remove noises), and then compute a rotation matrix $R_i^Z$ based on the normal of the segment-level ground points and the normal computed from the over-view point clouds to achieve segment level 3D registration; this is then followed by a Z-plane correction by computing the Z-offset $t_i^z$ between the street-view and over-view ground points.

With the resulting transformation parameters after the heuristic searching process, we obtain for each street-view segment, the 2D offset $t_i^{2d}$, 2D rotation $R_i^{2d}$, 3D rotation $R_i^Z$ and Z offset $t_i^z$. Therefore, the final 3D transformation $T_i^{3d} = [R_i^{3d}|t_i^{3d}]$ from the raw street-view dense point to the over-view point cloud for each street-view segment $i$ can be expressed as:

$$R_i^{3d} = \begin{bmatrix} R_i^{2d} & 0 \\ 0 & 1 \end{bmatrix} R_i^Z \qquad (5)$$
$$t_i^{3d} = [t_i^{2d}|t_i^z]$$

Although this process is heuristic while it practically has yield promising results, and an example is shown in **Figure 6** that depicts a building segment with and without the point cloud level refinement. This process will output a list of 3D transformations for extracted street-view building segments (as introduced in **Section 3.3.1**) and their associated point clouds $\mathcal{T}^{3d} = \{T_i^{3d}, i = 1, 2, \dots, M\}$ for subsequent processing.

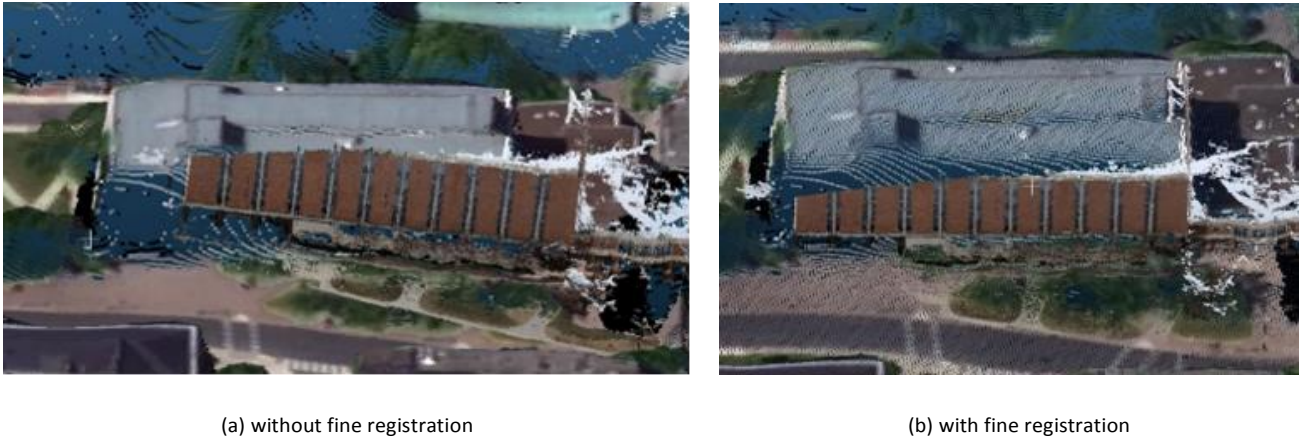

(a) without fine registration    (b) with fine registration

*Figure 6:* An example depicting the combined point clouds with and without the point-cloud level fine registration.

### 3.4 Bundle adjustment

With the registered street-view point clouds with the rigid transformations for each segment $\mathcal{T}^{3d} = \{T_i^{3d}, i = 1, 2, \ldots, M\}$, we can re-optimize the camera poses such that these are consistent with the registered street-view point clouds, so to benefit applications such as texture mapping. The re-optimization requires the 3D and 2D correspondences to be pre-recorded during the preprocessing stage, thus when the 3D points are transformed based on $\mathcal{T}^{3d}$, the correspondences can be reused as observations to readjust the poses. Here we assume the 3D and 2D correspondences are kept for the street-view SfM reconstruction, and we employ a robust transformation procedure using a distance-weighted averaging when applying the piece-wise rigid transformation $\mathcal{T}^{3d}$: for each 3D point belonging to a segment $i$, the transformation not only applies the $T_i^{3d}$, but also $T_j^{3d}$ (where $j \in \mathcal{N}_i$ are neighboring segments) to this point, and the transformed point takes a weighted average of 3D points produced by applying these transformations, where the weight is inversely proportional to the distance between the point and the segment $i$, as follows:

$$T^{3d} = \frac{\sum_{1 \leq i \leq M} w_i * T_i^{3d}}{\sum w_i} \tag{6}$$

$$w_i = 1/log(d_i)$$

where $d_i$ is the distance between a 3D point to the street-view building segment, and the similar transformation can be performed directly on the original poses of the image, by applying the appropriate $t_i^{3d}$ to the translation of the pose and the $R_i^{3d}$ on the rotation matrix, resulting in the registered image poses $\{(R_i, t_i)\}$.

To ensure the registered image poses follow the collinearity equation and the epipolar geometry, we re-optimize the poses through a constrained bundle adjustment, the constraint being that the estimated poses should be close to the registered pose while satisfy collinearity equations. Hereby the formulation is: for the registered street-view 3D points $\{X_j\}$ and the 2D observations $u_{i,j}$, as well as the registered pose of each image $\{(R_i, t_i)\}$ we minimize:

$$\min_{\{X_j\},\{(R_i,t_i)\}} \sum_{i,j} \|u_{i,j} - f(X_j|R_i, t_i)\|_2^2 + \sum_i \lambda \|\Delta t_i\|_2^2 \tag{7}$$

where $f(X_j|R_i, t_i)$ is the projection of a 3D point $X_j$ onto the image $\{(R_i, t_i)\}$ following the collinearity equations (Thompson et al., 1966) and $u_{i,j}$ is the corresponding feature point on the image, $\Delta t_i$ is difference between the estimated and the registered pose. $\lambda$ is a constant parameter to leverage the contribution of the constraint and it is set to 20 in our experiments. The first term of this equation minimizes the reprojection error and the second penalize the poses to be too different from the registered poses. The constrained bundle adjustment problem in **Equation 7** can be easily implemented and solved using the Ceres Solver (Agarwal and Mierle, 2012) and the yielded image poses can be used to optionally to reproduce the dense street-view point clouds resulting in globally consistent data with respect to the over-view point clouds. This process of re-optimizing the image poses and reproducing the dense point clouds using the new poses, may correct minor registration errors of individual building segments in previous processing.

## 4 Experiments

The testing region is in the main campus of The Ohio State University in Columbus, Ohio, USA. covering an area of ca. 16 $km^2$. Twelve WorldView I/II satellite images covering the testing region are collected and processed using the approach described in Qin et al. (2019b) based on the satellite stereo processing software RSP (Qin, 2017, 2016) to generate the DSM, orthophoto and point clouds for this region. For the street-view data, we mounted a GoPro Hero 7 camera on the top of a car and drive around the campus covering a trajectory equivalent to 33 km (resulting in approximately 300GB of video data). The speed of the vehicle is 25 mph on average, which was used to estimate the scale $s$.

We quantitatively analyzed the accuracy of the proposed cross-view geo-registration algorithm in both 3D and 2D. For 3D analyses, a Lidar point cloud is collected from a mobile platform covering a 1.5 km street in length in our testing region, data shown in **Figure 7(a).** The 2D analyses evaluates the proximity of the projected building boundaries from street-view to the derived building boundaries of the over-view DSM. **Figure 7(b)** shows the DSM building boundaries of an area as the reference (manually edited to ensure the fidelity of the evaluation). The registered street-view dense point clouds (reproduced using the optimized pose described in **Section 3.3.3**) is shown in **Figure 7(c)**, which present no visible trajectory drift as compared to the original street-view point clouds directly computed through SfM (**Figure 3(a)**).

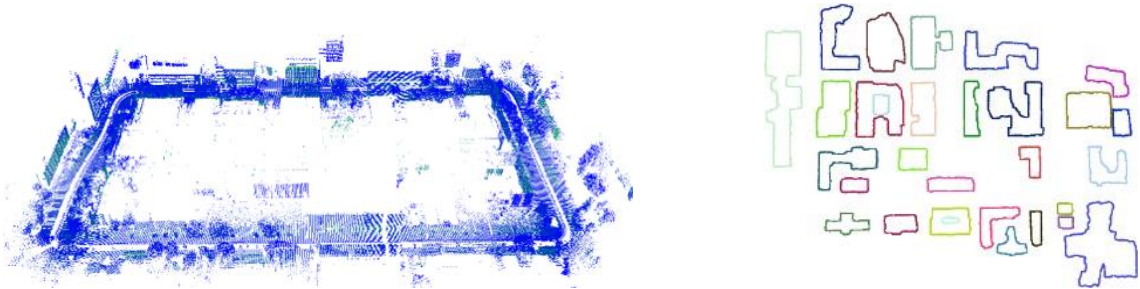

(a) LiDAR point clouds used for 3D evaluation    (b) 2D reference DSM boundaries used for evaluation

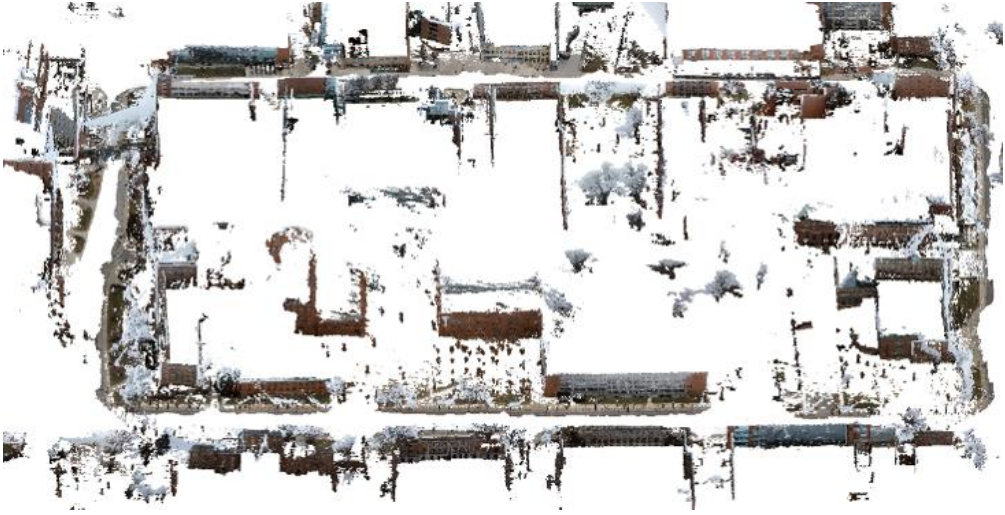

(c) registered street-view points

*Figure 7:* The 3D Lidar point clouds (a) and 2D DSM building boundaries (b) are utilized as ground truth for the accuracy evaluation of the registered street-view dense point clouds (c) in our experiment. The length for this close looped area is 1.5 *km*, and buildings on both sides are scanned. Segments shown in (b) colorized to differentiate individual segments.

### 4.1 Registration accuracy

The accuracy of the geo-registration in 3D is evaluated by comparing the street-view point clouds (**Figure 7(c)**) to the LiDAR-based 3D scan (**Figure 7(a)**). The symmetric cloud-to-cloud distance metric - the Chamfer distance is used for evaluation: here we compute for each point $p_i^{3d}$ in the street-view point clouds $P_f^{3d}$ its closest distance to the LiDAR point clouds $P_l^{3d}$ as $d_c(p_i^{3d}, P_l^{3d})$ symmetrically for each LiDAR point $p_j^{3d}$ in $P_l^{3d}$, its closest distance to the street-view point clouds as $d_c(p_j^{3d}, P_f^{3d})$. Here we only consider points are symmetrically consistent in these two clouds to form the Chamfer distance: $p_i^{3d}$ from $P_f^{3d}$ and $p_j^{3d}$ from $P_l^{3d}$ are mutually the closest point to each other:

$$\begin{cases} d(p_i^{3d}, p_j^{3d}) = d_c(p_i^{3d}, P_l^{3d}) \\ d(p_j^{3d}, p_i^{3d}) = d_c(p_j^{3d}, P_f^{3d}) \end{cases} \tag{8}$$

where $d(\cdot,\cdot)$ refers to 3D Euclidean distance. The symmetric Chamfer distance averages the distances of these points as the cloud-to-cloud distance. To accelerate and remove outliers of the point clouds, we disregard any point-to-point distance that are larger than 10 m, which has yielded the 1.27 meters of error and its point level distribution is shown in **Figure 8(a)**, which can be seen that most of the per point errors are less than 2 meters and the computed standard deviation (std.) of this error distribution is 1.34 meters. Both the metrics (Chamfer distance and the std.) at the range of 1-1.5 meters indicates the registration has matched to the over-view dataset at the level of 2-3 pixels considering its GSD being 0.5 meters.

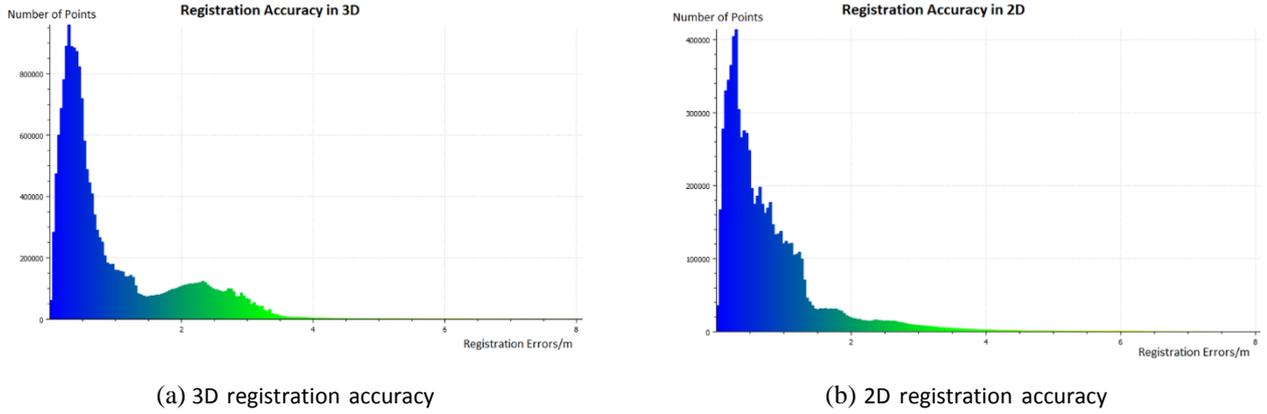

(a) 3D registration accuracy  (b) 2D registration accuracy

*Figure 8*: Registration accuracy in 3D and 2D by comparing to the ground-truth Lidar point cloud and the over-view buildingboundary points. For the 3D case, the average and mean squared errors are 1.27m and 1.34m, respectively. For the 2D case, those values are 0.99m and 1.04m, respectively.

The similar evaluation is performed at the 2D level, which uses the same distance metric (simply reduced to 2D), and gives a horizontal metric accuracy (in Chamfer distance) of 0.99 meters with a standard deviation of 1.04, with distribution show in **Figure 8(b)** indicating errors are mostly less than 2 meters. This is equivalent to two pixels of accuracy in the over-view data.

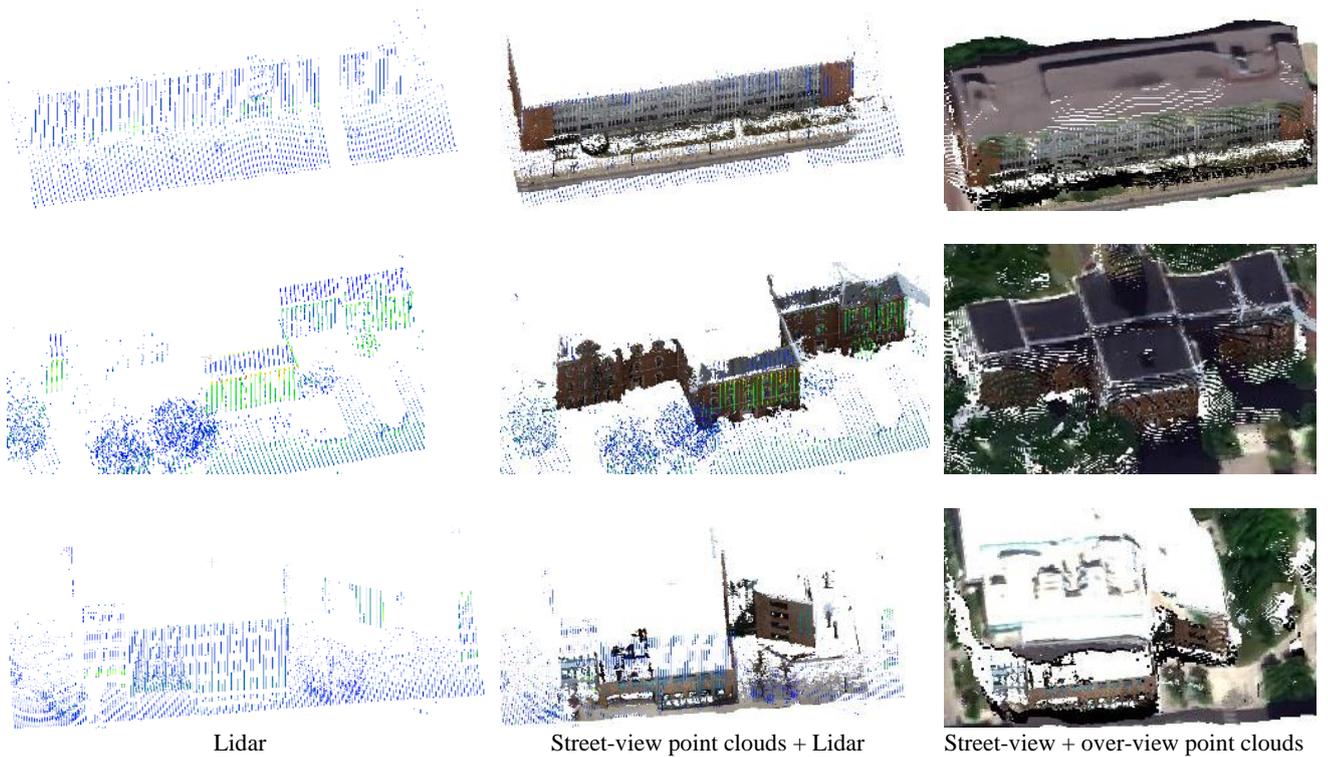

Lidar    Street-view point clouds + Lidar    Street-view + over-view point clouds

*Figure 9:* Examples of registered street-view point clouds to the over-view point clouds. From left to right: LiDAR point cloud, the street-view point clouds as compared to LiDAR, combined street-view and over-view point clouds.

**Figure 9** shows a few detailed results at the building level for visual assessment, LiDAR point clouds drawn to serve as a reference. It can be seen that these façade points (from the street-view) and the roof points are consistently aligned well and form the complete point clouds for the buildings.

### 4.2  Registration of street-view point clouds to OpenStreetMap and semantic labels

Since our proposed methods take the building boundaries/segments as the major source for registration, it can be flexibly applied to any kind of 2D vector data. Here we assume only 2D over-view information are available, in the form of OpenStreetMap (OSM) (OpenStreetMap, 2021) building segments, or segments detected from the Orthophoto through semantic segmentation. An example of these two types of data are shown in **Figure 10(a)** and **(b)**, respectively the OSM data and the semantic labels detected using a U-Net detector described in (Qin et al., 2019a,b). We simply regard these two types of data as our over-view building segments as apply the algorithms described in **Section 3.3.1-3.3.2**. Note for the OSM, we intentionally eliminated building segments that are under 5 meters in diameter, to match the detectable buildings in satellite data for fair comparison. Results of the registration in a small test region is shown in **Figure 10(c)** and **(d)**, which shows that the proposed method works equivalently well for different types of data. The Chamfer distance and the std. of the error (as used for accuracy evaluation in **Section 4.1**) are 1.21 meters & 1.34 meters for OSM data, and 1.46 meters & 1.59 meters for semantic labeling dataset, which similarly match the accuracy concluded in **Section 4.1**. Note that all these experiments were performed using the same parameters throughout this work without tuning.

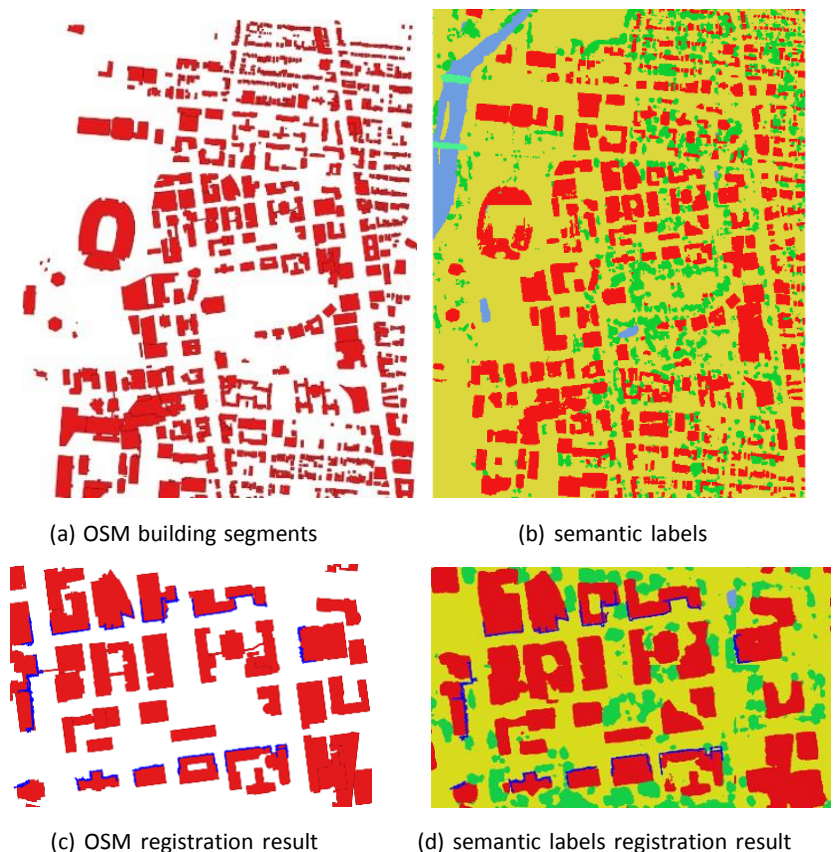

(a) OSM building segments      (b) semantic labels

(c) OSM registration result      (d) semantic labels registration result

*Figure 10*: Registration results on the OSM and semantic labels dataset. Building segments from street-view are colored in blue. The Chamfer distance and the std. of the error are 1.21 meters & 1.34 meters for OSM data, and 1.46 meters & 1.59 meters for semantic labeling dataset.

### 4.3 Large-scale experiments

We further expand our evaluation scope by applying the proposed method to larger region that includes 10 trajectories of data involving approximately 150 K street-view images, covering an area of 16 $km^2$, with the cumulated trajectory 33 $km$ in length. The dataset has yielded over 1 billion street-view points and around 10 million over-view points for registration. Results of this region is shown in **Figure 11(a)** and **(b)**, respectively showing the street-view point clouds and the combined street-view and over-view point clouds, particularly in **Figure 11(b)** the large-scale point clouds shows that both the street-view and over-view point clouds are well aligned. Since there is no mobile LiDAR data available this experimental region, we only evaluate the 2D accuracy based on the over-view building segments as the ground truth. An average 2D registration error (Chamfer distance as described in **Section 4.1** of 1.24 m is obtained for the 10 trajectories. If ranked based on the registration accuracy, the top 5 trajectories with most buildings in this area has obtained an average 2D registration error of 1.05 m, while the rest obtained 1.52 m on average. **Figure 12(a)** and **(b)** show two examples where both point clouds are registered extremely well (with the registration errors are less than 0.5m, i.e. less than a pixel).

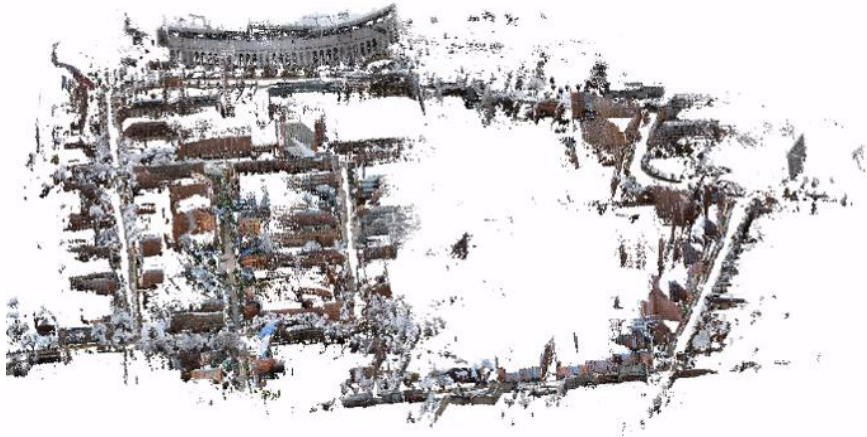

(a) street-view points

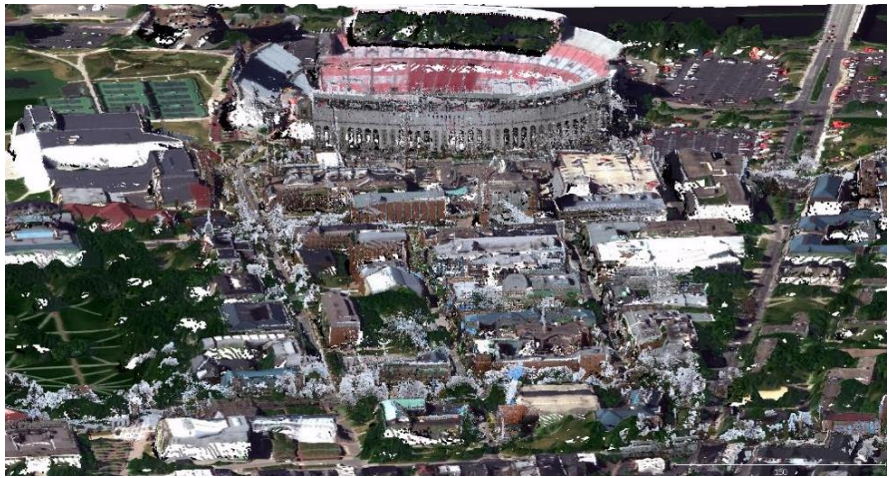

(b) combined street-view + over-view point clouds

*Figure 11:* Registration results in a large region covering 16 $km^2$. (a) is the registered street-view point clouds and (b) is the merged cross-view point clouds.

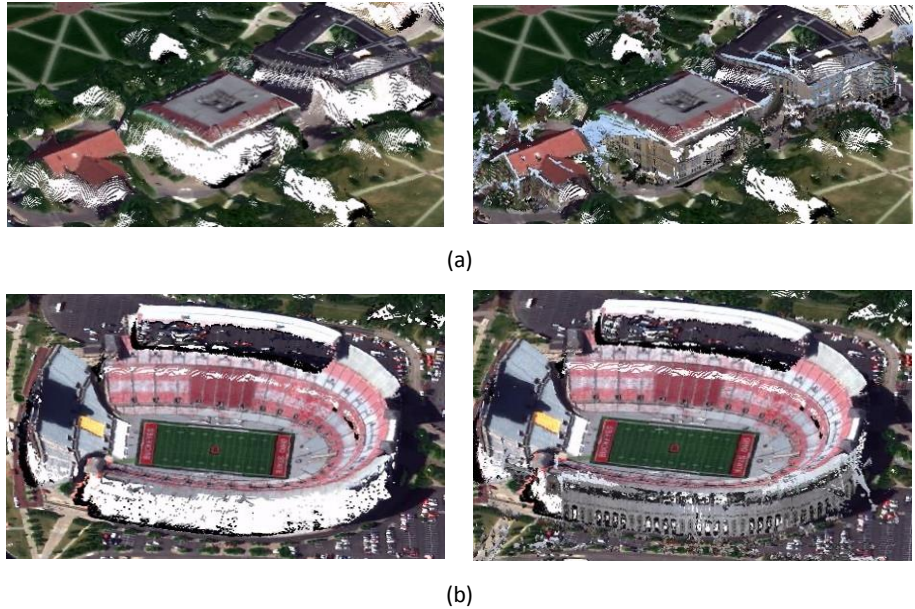

*Figure 12*: Two well-registered examples. (a) and (b) are two examples showing buildings with good street-view and over-view alignment (registration error less than 0.5 *m*). The over-view point clouds are in the first column while the cross-view pointclouds are in second column.

## 5 Conclusions

In this paper, we present a solution to address a novel and challenging problem in 3D data registration, where we aim to automatically register the street-view point clouds generated from monocular video cameras, to over-view point clouds generated from multi-view satellite images (with 0.5 meters). The problem is challenging in that 1) the over-view and street-view point clouds are compensatory in terms of their line of sight, and share very little common area for registration; 2) there exist a huge difference in terms of the resolution of two types of data (a factor of a hundred); 3) the street-view trajectory presents non-rigid, non-parametric topographical distortions that are not suitable for simple and parametric registration models (e.g. rigid transformation). To this end, we observe that the most informative information to tie these two datasets are building boundaries which are extractable from both of the two datasets, thus the solution we developed evolved around this observation by firstly perform 2D building boundary matching, where challenges of matching incomplete building segments are tackled using a global optimization framework, and the 3D registration are further refined through a heuristically exhaustive search. The camera poses are then readjusted based on the registration results to generate consistent and well-registered street-view point clouds. The proposed method is evaluated with a rich dataset that involves as many as ten trajectories of video frame data (150 K in total) with 33 $km$ in length, covering an region of 16 $km^2$, we show that our proposed method achieves 0.99 m registration accuracy in horizontal direction, a 1.27 m in 3D; we have also demonstrated that the proposed method is flexible enough to register to other type of dataset including OSM and building segments extracted from semantic segmentation results. The proposed method assumes the scale of the street-view data is known, while it is designed to be GPS agonistic and does not require any initial correspondences between the over-view and street-view data. In occasions where coarse correspondences are known, they can be easily adapted to our proposed framework. However, since the proposed method heavily depends on boundaries of buildings or off-terrain objects, we found that in practice that the registration accuracy can be dependent on the building boundaries, and apparently in scene with very sparse buildings/off-terrain objects, the proposed method may fail, in which coarse GPS information might be used to improve the robustness of the algorithm, and this will be attempted in our future work.

**Disclaimer:** Mention of brand names in this paper does not constitute an endorsement by the authors.

# 6 Acknowledgement

The study is supported by the ONR grant (Award No. N000141712928). We would like to thank Prof. Charles Toth for providing the LiDAR dataset for the OSU campus and acknowledge Mr. Xiaohu Lu's prior assistant to the work.